\documentclass[10pt,twocolumn,letterpaper]{article}

\usepackage{iccv}
\usepackage{times}
\usepackage{epsfig}
\usepackage{graphicx}
\usepackage{amsmath}
\usepackage{amssymb}
\usepackage{bbm}
\usepackage{booktabs}
\usepackage{multirow}
\usepackage{bbding}
\usepackage{subfigure}
\usepackage{amsthm}
\usepackage{makecell}
\usepackage{mathrsfs}
\usepackage[dvipsnames]{xcolor}
 \newcommand{\compresslist}{%
 \setlength{\itemsep}{0pt}%
 \setlength{\parskip}{0pt}%
 \setlength{\parsep}{0pt}%
 }

\usepackage[pagebackref=true,breaklinks=true,letterpaper=true,colorlinks,bookmarks=false]{hyperref}

\hypersetup{
    colorlinks,
    citecolor=CornflowerBlue,
}

\def\iccvPaperID{***} 

\iccvfinalcopy
\ificcvfinal\fi

\begin{document}

\title{PlaneTR: Structure-Guided Transformers for 3D Plane Recovery} 

\author{Bin Tan\footnotemark[1]$~~^1$ \quad Nan Xue\footnotemark[1]$~~^1$
\quad Song Bai$~^2$ \quad Tianfu Wu$~^3$ \quad Gui-Song Xia\footnotemark[2]$~~{}^1$\\
$^1$ School of Computer Science, Wuhan University\\
$^2$ ByteDance AI Lab \quad 
$^3$ Department of ECE, NC State University\\
{\fontsize{10}{10}\selectfont \url{https://git.io/PlaneTR} 
}
}

\maketitle

\renewcommand{\thefootnote}{\fnsymbol{footnote}}
\footnotetext[1]{Equal Contribution.} 
\footnotetext[2]{Correspondence Author.}
\ificcvfinal\thispagestyle{empty}\fi

\begin{abstract}
This paper presents a neural network built upon Transformers, namely PlaneTR, to simultaneously detect and reconstruct planes from a single image. Different from previous methods, PlaneTR jointly leverages the context information and the geometric structures in a sequence-to-sequence way to holistically detect plane instances in one forward pass. Specifically, we represent the geometric structures as line segments and conduct the network with three main components: (i) context and line segments encoders, (ii) a structure-guided plane decoder, (iii) a pixel-wise plane embedding decoder. Given an image and its detected line segments, PlaneTR generates the context and line segment sequences via two specially designed encoders and then feeds them into a Transformers-based decoder to directly predict a sequence of plane instances by simultaneously considering the context and global structure cues. Finally, the pixel-wise embeddings are computed to assign each pixel to one predicted plane instance which is nearest to it in embedding space. Comprehensive experiments demonstrate that PlaneTR achieves a state-of-the-art performance on the ScanNet and NYUv2 datasets.
\end{abstract}


\section{Introduction}
Recovering 3D planar structures from a single RGB image is a fundamental problem in 3D vision and is challenging due to its ill-posed nature. The goal of this problem is to detect regions of plane instances and estimate their 3D planar parameters (e.g. surface normal and offset) in an image. As a fundamental representation of 3D scenes, the reconstructed planes have a wide range of applications in downstream tasks, such as augmented reality \cite{chekhlov2007ninja}, visual SLAM \cite{taguchi2013point, yang2019monocular, ma2016cpa} and indoor scene understanding \cite{tsai2011real, li2020multi}.

\begin{figure}
\centering
\includegraphics[width=0.95\linewidth]{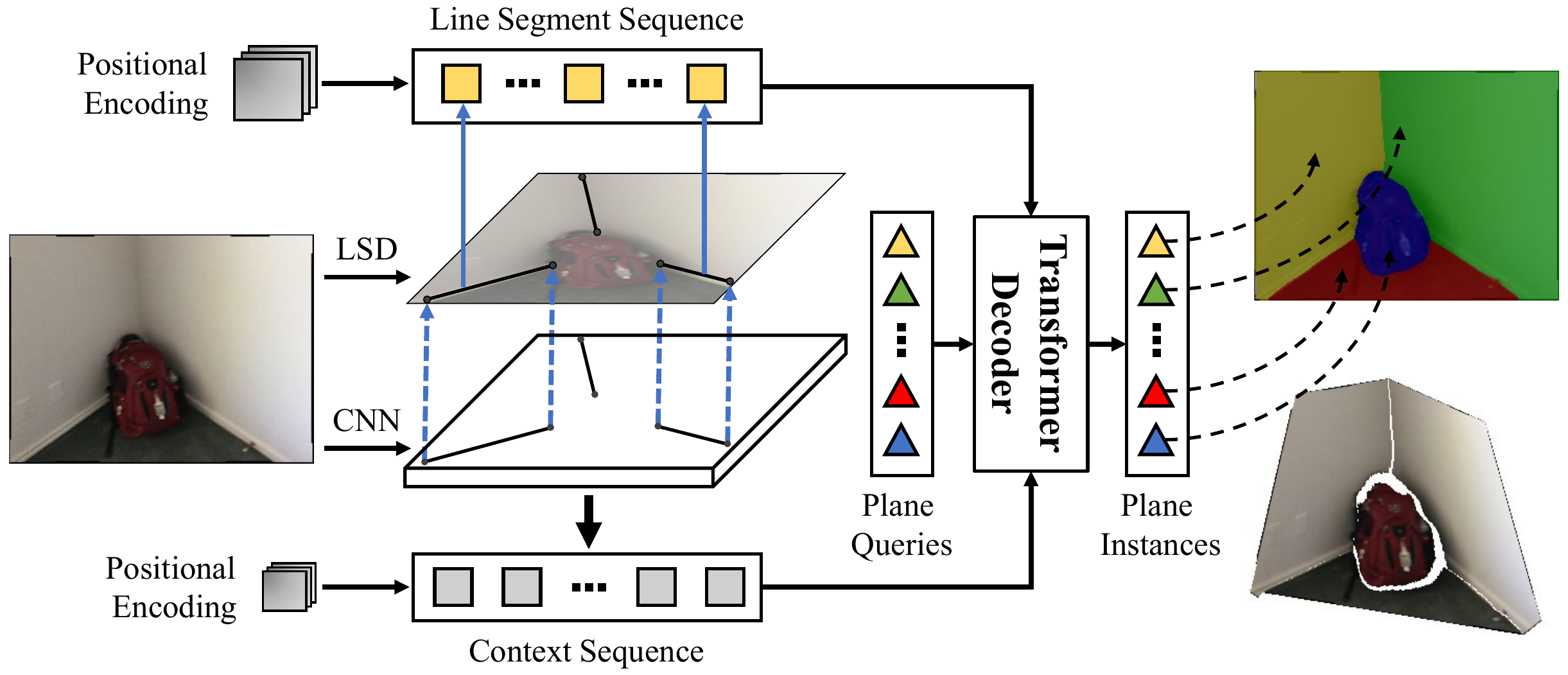}
\caption{Illustration of the proposed PlaneTR. Our network holistically leverages context features and line segments represented by two tokenized sequences to predict a set of plane instances in an image.}
\label{fig:network}
\end{figure}

Some early methods \cite{delage2007automatic, barinova2008fast, micusik2008towards, GeometricRecovery, qian2018ls3d} tend to utilize geometric elements such as line segments, junctions, and vanishing points to tackle this problem in a bottom-up manner. These geometric elements are usually first divided into various groups and then carefully analyzed under a series of strict presuppositions (e.g. Manhattan World) and rules to recover 3D planes. Although these structure-based methods have achieved successes to some extent, they are suffering from the issues of missing or incorrect detection of geometric primitives, complex technique process, and limited scenes which affect their performance and applications.

Recently, Convolutional Neural Networks (CNNs) addressed this problem  \cite{planeNet, PlaneRecover, planercnn, PlaneAE19, Jiang2020Peek, interPlane}. Some methods \cite{planeNet, PlaneRecover, planercnn} directly predict plane instance masks with corresponding 3D planar parameters from the input image in a top-down manner. By contrast, PlaneAE \cite{PlaneAE19} takes a bottom-up manner and achieves plane instances by clustering pixels that are similar in an embedding space. These methods relax constraints in structure-based methods and have achieved promising performance. However, they mainly leverage the context information from CNNs and ignore structure cues in the image which are useful for 3D plane recovery.

In this paper, we are interested in exploiting geometric structures for the problem of 3D plane recovery of the indoor scene under a learning-based framework. Although there are various low-level geometric primitives, we find that line segments are usually used to construct 3D planes \cite{li2020multi, qian2018ls3d} and contain more holistic 3D information of the scene when comparing with other geometric primitives, such as feature points, edges, and vanishing points. Besides, benefiting from recent state-of-the-art works in line segment detection \cite{xue2020holistically,xue2019AFM,xue2019AFMpami,zhou2019lcnn}, it is convenient for us to achieve line segments from an image. Thus, in this paper, we use line segments as the geometric structures for plane recovery.


In some recent works \cite{wang2020vplnet, jin2020geometric}, geometric structures have already been used for depth and normal estimation. In these methods, structures are represented as dense maps (e.g. line segment map) to meet the representation of CNNs. However, such dense representation of structures is hard for the network to leverage global structure cues with two drawbacks: (i) limited receptive field of CNNs and (ii) sparse distribution of structure pixels in the dense map. Although CNNs-based attention mechanisms \cite{wang2018nonlocal,woo2018cbam} can alleviate the first drawback, the second one is still hard to be well tackled. Therefore, this paper is going toward answering the question: 
\begin{quote}
\emph{
    If it is possible to holistically exploit sparse line segments for the learning of plane recovery?}
\end{quote}

Most recently, the sequence-to-sequence model of Transformers \cite{vaswani2017attention} has been successfully used in vision tasks \cite{carion2020DETR, xu2021line}. In these works, the input features and output targets are represented as visual tokens and globally interacted with each other via the attention mechanism of Transformers. Motivated by the tokenized representation in vision Transformers, we address the above question by proposing our PlaneTR, a Transformer model that leverages the informative context features and meaningful geometric structures for plane recovery. 

For a given input image and its detected line segments, our PlaneTR encodes the line segments and context features into two sets of tokenized sequences, respectively. Then, a set of learnable plane queries are used to holistically interacted with the context and line segment sequences via a structure-guided plane decoder which outputs a set of tokenized plane instances. As a final step, we design a simple instance-to-pixel segmentation strategy motivated by the associative embedding~\cite{Newell2017Emb} and PlaneAE~\cite{PlaneAE19}, which yields the pixel-wise plane segmentation results by assigning each pixel to its nearest plane instance in the embedding space.

In summary, the main contributions of this paper are as follows:
\begin{itemize}\compresslist
    \item
    We leverage line segments as tokenized sequences instead of dense maps to guide the learning of 3D plane recovery with geometric structures. 
    
    \item
    We develop a novel Transformer, PlaneTR, to simultaneously detect and reconstruct plane structures from a single image in a sequence-to-sequence manner.

    \item
    Our method obtains a new state-of-the-art on the ScanNet \cite{dai2017scannet} and NYUv2 \cite{silberman2012indoor} datasets, verifying the effectiveness of our method.
\end{itemize}

\section{Related Work}
\subsection{Single Image 3D Plane Recovery}
Geometric structures such as line segments and vanishing points are widely used in traditional methods \cite{delage2007automatic, barinova2008fast, micusik2008towards, GeometricRecovery, qian2018ls3d, Langlois2019Surface, zaheer2018single, fouhey2014unfolding} for piece-wise 3D plane recovery from a single image. These methods usually tend to leverage structures via a grouping and optimization (or fitting) strategy to solve this problem. For example, Lee \etal~\cite{GeometricRecovery} propose to reconstruct the planar 3D scene from a single image by first proposing a number of structure hypotheses and then finding a model which best fits the collected line segments. Qian \etal~\cite{qian2018ls3d} design a Line-Segment-to-3D algorithm which first groups line segments into a set of minimal spanning Manhattan trees and then lifts these trees to 3D under the Manhattan constraints. Li \etal~\cite{li2020multi} assume a box prior to the scene and propose to find the best representation of the image via a structure-guided search algorithm. To effectively conduct their algorithms, these methods are generally built upon a series of strict assumptions and rules which limited their application and performance.

The recent learning-based methods \cite{PlaneRecover,planeNet,PlaneAE19, planercnn, interPlane} relax constraints in traditional structure-based methods by directly detecting and reconstructing 3D planes from single-view images. PlaneNet \cite{planeNet} generates a large planar depth maps dataset from the ScanNet dataset \cite{dai2017scannet} and proposes an end-to-end deep neural network to predict a set of plane segmentation masks and their corresponding 3D planar parameters. PlaneRecover \cite{PlaneRecover} designs a plane structure-induced loss and trains a plane reconstruction network directly from the RGB-D data without 3D plane annotations. PlaneRCNN \cite{planercnn} addresses the plane recovery problem by applying a detection-based framework (e.g. Mask R-CNN \cite{he2017mask}) and proposes a segmentation refinement network and a warping loss to improve performance. Different from previous top-down methods, PlaneAE \cite{PlaneAE19} solves this problem in a bottom-up manner. It obtains plane segmentation masks by first learning an embedding for each pixel and then grouping them to various plane instances via an efficient mean shift algorithm. Despite the promising performance achieved by these learning-based methods, they only consider the context information in deep features and ignore structure cues which are important for 3D plane recovery. By contrast, we combine the idea from both traditional and learning-based methods and propose a method to jointly utilize the context information and geometric structures for 3D plane recovery.

\begin{figure*}
\centering
\includegraphics[scale=0.5]{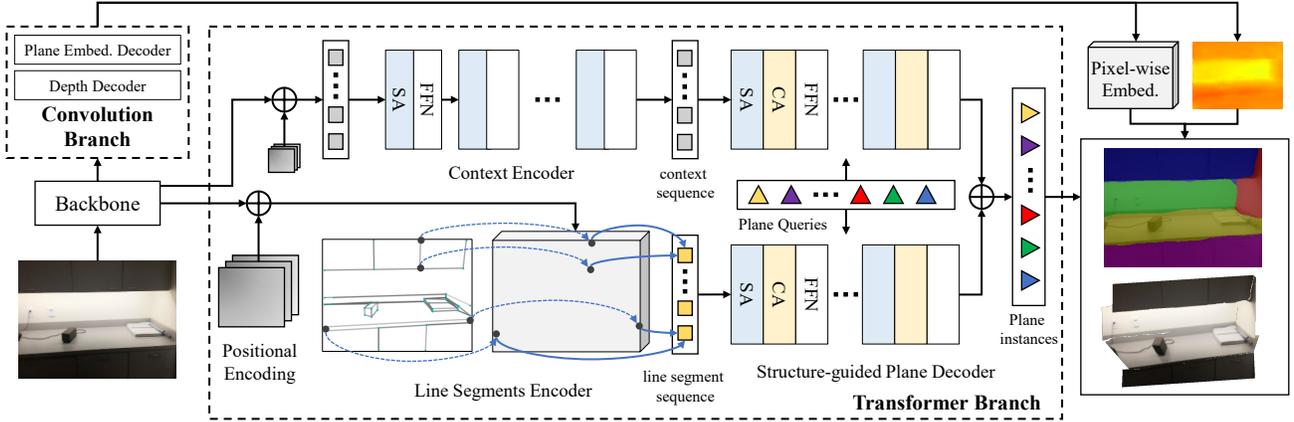}
\caption{Overview of our proposed PlaneTR. It mainly consists of two branches: (i) a Transformer branch which models holistic relations between plane instances and context (line segment) sequence; and (ii) a convolution branch which predicts pixel-wise embeddings used for plane segmentation and depths of non-plane regions.}
\label{fig:network_full}
\end{figure*}

\subsection{Structure Guided Learning}
Structures have been successfully used in some learning-based methods \cite{wang2020vplnet, jin2020geometric, ramamonjisoa2019sharpnet, song2020edgestereo, qiu2020pixel}. Wang \etal~\cite{wang2020vplnet} propose to learn a Manhattan Label Map from the input RGB image and its corresponding Manhattan line map for normal estimation. Jin \etal~\cite{jin2020geometric} leverage geometric structure as a prior and a regularizer to facilitate the learning of depth map. In their method, they first estimate geometric structures (e.g. corner map, boundary map and plane map) from the input image as a prior and then estimate the same structures from the output depth map as a regularizer. Song \etal~\cite{song2020edgestereo} propose to embed learned edge features into the deep context features to guide the estimation of disparities in image details and boundaries. These methods tend to serve structures as local guidance and represent them as pixel-level dense maps which are difficult for the network to utilize global structure cues. Different from these methods, in this paper, we take a novel tokenized representation of line segments to introduce global structure cues into a learning-based framework. 

\subsection{Vision Transformers}
Transformers \cite{vaswani2017attention} are proposed for sequence-to-sequence machine translation in NLP. Recently, some researchers attempt to introduce Transformers into vision tasks, such as objection detection \cite{carion2020DETR,zhu2020deformableDETR}, image segmentation \cite{wang2020maxDeepLabTR,wang2020VISTR,liang2020polytransformTR}, and image classification \cite{dosovitskiy2020imageViT}. DETR \cite{carion2020DETR} proposes a new simple paradigm for object detection built upon a Transformer encoder-decoder architecture which is free from many hand-designed components such as anchor generation and non-maximum suppression. DETR formulated the problem of object detection into a sequence-to-sequence prediction problem and directly predicts a set of objects from the learned object queries which are interacted with the context feature sequence. Liang \etal~\cite{liang2020polytransformTR} exploit a self-attention network in Transformers to model the intricate dependencies in polygon vertexes and learn to predict a set of offsets for the input vertexes to deform the initial polygon to object boundaries. Inspired by DETR, Xu \etal~\cite{xu2021line} propose a network to predict a set of line segments directly from tokenized image features which simplifies the line segment detection process and achieves state-of-the-art performance. On one hand, these works show the promising application of Transformers in vision tasks. On the other hand, such sequence-to-sequence architecture based on the tokenized representation gives us a suitable manner to jointly leverage the instance-level structures and context features for plane recovery. Thus, in this paper, we propose a novel structure-guided network built upon Transformers for the 3D plane detection and reconstruction, namely PlaneTR. 

\section{The PlaneTR Model}

As illustrated in Fig.~\ref{fig:network_full}, the proposed PlaneTR is made up of a convolutional backbone followed by a Transformer branch for plane instances prediction (sec.~\ref{sec:PPM}) and a convolution branch for pixel-wise plane embedding and non-plane region depth estimation via two decoders. The estimated pixel-wise embeddings are then used to plane instance segmentation (Sec.~\ref{sec:PlaneSeg}). We use the modified HRNet-w32 \cite{WangSCJDZLMTWLX19} as our backbone which takes an image $I\in \mathbb{R}^{H\times W\times3}$ as input and outputs feature maps with four scales, denoted by $\mathbb{F}=\{F_{i}\in \mathbb{R}^{H_i\times W_i\times C_{i}}\}_{i=1}^{4}$ ($H_i=H/2^i$, $W_i=W/2^i$). According to \cite{PlaneAE19}, we define the 3D parameter of a plane as $n\doteq \tilde{n}/d \in \mathbb{R}^3$, where $\tilde{n} \in \mathbb{R}^3$ is the surface normal and $d$ indicates the distance from plane to camera center.

\subsection{Plane Instance Prediction with Transformers}
\label{sec:PPM}


\paragraph{Context Encoder.} We first encode context features from the backbone feature map $F_4$ into a tokenized feature sequence with a standard Transformer encoder which consists of six encoder layers as used in DETR~\cite{carion2020DETR}. Specifically, we feed a flattened feature sequence $f_c\in \mathbb{R}^{d\times (H_4 W_4)}$ generated from $F_4$ followed by a $1 \times 1$ convolution and its sine/cosine positional encoding $E_c\in \mathbb{R}^{d\times (H_4 W_4)}$ into the encoder. In each encoder layer, the input feature tokens are interacted with each other via a self-attention (SA) operation, and a fully connected feed forward network (FFN) is used to get the output. The final output context sequence is defined as $S_c\in \mathbb{R}^{d\times (H_4 W_4)}$. Here, $d$ is set to 256 in this paper.

\vspace{-10pt}
\paragraph{Line Segments Encoder.} To achieve the tokenized line segment sequence, we first detect $n$ line segments $\mathbb{L}=\left\{ \mathbf{l}_i = (\mathbf{x}_i^1, \mathbf{x}_i^2 ) \right\}_{i=1}^{n}$ using the state-of-the-art line segment detection algorithm HAWP \cite{xue2020holistically} with their pretrained model. 
Then, we build the line segment sequence based on a high-resolution backbone feature $F_2$ in order to distinguish adjacent line segments.

Denoted by ${F_r}\in \mathbb{R}^{H_2 \times W_2 \times d}$, a $1\times 1$ convolution layer is applied on the top of the backbone feature map $F_2$. After computing the positional encoding map $E_r$ of $F_r$, we sample feature vectors and the positional encoding vectors for each line segment $\mathbf{l}$ at its two endpoints using the bi-linear interpolation, denoted by $(f_1,f_2)$ and $(e_1,e_2)$, respectively. In the next step, an MLP layer is applied to yield the feature of line segment $\mathbf{l}$ and its corresponding positional encoding as:
\begin{equation}
\label{eq: line_feat}
f_{\mathbf{l}} = \text{MLP} \left(f_{1} \otimes f_{2}\right), e_{\mathbf{l}} = (e_1 + e_2) / 2,
\end{equation}
where $\otimes$ indicates a concatenation operation. The features and the positional encodings of all line segments are concatenated into $\hat{S}_l, E_l\in\mathbb{R}^{d\times n}$ as the initial line segment sequence. Finally, another MLP layer is applied to $\hat{S}_l$ to output the final feature sequence of line segments, denoted by  $S_l \in \mathbb{R}^{d\times n}$.
\vspace{-10pt}
\paragraph{Structure-guided Plane Decoder.} Our Structure-guided Plane Decoder has two paralleled branches as shown in Fig.~\ref{fig:network_full}. Each branch is built upon a standard Transformer decoder which consists of six encoder layers as used in DETR~\cite{carion2020DETR}. The two branches take the context and the line segment sequences as inputs respectively, and share the same learnable plane queries $E_p\in \mathbb{R}^{d \times K}$. Here, $K$ is a fixed number of predicted planes ($K=20$ in this paper). By interacting with the context and line segment sequence simultaneously with a cross-attention (CA) operation of the Transformer, the plane queries are able to perceive plane instances by holistically consider the context and structure cues in an image. We define the output sequences of the two branches as $O_c$ and $O_l$, respectively. Then, the final decoded plane instance sequence of the Structure-guided Plane Decoder can be calculated as $S_p = O_c + O_l\in \mathbb{R}^{d \times K}$. 


\vspace{-10pt}
\paragraph{Plane Instance Prediction Layer.} For each plane instance in $S_p$, we apply a multi-head linear layer to predict a set of parameters for the plane instance, including the 3D planar parameters $n\in \mathbb{R}^{3}$, the plane and non-plane probability $p$ and the plane instance embedding $\mathcal{E} \in \mathbb{R}^{\epsilon}$ ($\epsilon=8$ in this paper), respectively. In the inference stage, we select the predicted plane instances as outputs if their plane probabilities are larger than 0.5.

\subsection{Instance-to-Pixel Plane Segmentation}
\label{sec:PlaneSeg}
After achieving plane instances from the Transformer branch, it is required to obtain the segmentation mask for each plane. 
Inspired by PlaneAE \cite{PlaneAE19}, we apply the idea of associative embedding and calculate pixel-wise embedding vectors via a Plane Embedding Decoder as used in PlaneAE in the convolution branch. However, instead of achieving plane instance segmentation masks via the clustering algorithm based on pixel embedding, we take a simple instance-to-pixel method by directly comparing the distance between pixel embedding and plane instance embeddings. Then, we assign a pixel to one plane instance which is nearest to it in the embedding space if their distance is lower than a threshold $T$ which is set to 1.0 in this paper.

\subsection{Network Training}
\label{sec:training}
\paragraph{Auxiliary Plane Center Prediction.} We find that predicting auxiliary plane centers in the training stage is beneficial (we will discuss this point in the experimental ablation study).
The ground truth plane center of each plane is defined as the normalized average image coordinate of its corresponding plane pixels. In practice, we predict plane centers for both plane instances and pixels. The instance-wise plane centers $c \in \mathbb{R}^{2\times K}$ are predicted from the outputs of Structure-guided Plane Decoder followed by a linear layer. The pixel-wise plane centers $\mathcal{C} \in \mathbb{R}^{H\times W\times 2}$ are predicted via a convolution decoder which has the same architecture as the Plane Embedding Decoder.

\vspace{-10pt}
\paragraph{Bipartite Matching.} In our network, we predict a fixed number of $K$ plane instances $\{ s_p^i = (p_i, n_i, c_i, {\mathcal{E}}_i) \}_{i=1}^{K}$, which are generally more than the ground truth plane instances $\{ {\hat{s}}_p^j = ({\hat{p}}_j, {\hat{n}}_j, {\hat{c}}_j) \}_{j=1}^{M}$ in the image ($K \ge M$). Here, ${\hat{p}}_j\in \{ 0, 1\}$ (0: non-plane, 1: plane). Therefore, to effectively train the network, we first pad the ground truth plane instances to the number $K$ with non-plane instances. Then, we find a bipartite matching between the predict plane instances and the ground truth plane instances via searching for a permutation $\hat{\sigma}$ which has a minimization matching cost as:
\begin{equation}
\label{eq: bi-matching}
\hat{\sigma}=\underset{\sigma}{\arg \min } \sum_{i=1}^{K} D\left(\hat{s}_{p}^{i}, s_p^{\sigma(i)}\right), 
\end{equation}
where $\sigma(i)$ indicates the matched index of the predict plane instance to the ground truth plane instance $\hat{s}_{p}^{i}$ and $D$ is a function that measures the matching cost between two plane instances. The formulation of $D$ is defined as:
\begin{align}
\label{eq: bi_matching2}
D=-p_{\sigma(i)}\left( \hat{p}_{i}\right)
&+ \mathbbm{1}_{\{\hat{p}_{i}=1\}}  \,  L_{1}\left(\hat{{n}}_{i}, {n}_{\sigma(i)}\right) \nonumber\\
&+ \mathbbm{1}_{\{\hat{p}_{i}=1\}}  \, \omega \, L_{2}\left(\hat{{c}}_{i}, {c}_{\sigma(i)}\right),
\end{align}
where $\mathbbm{1}_{\mathbf{a}}$ is an indicator function taking $1$ if $\mathbf{a}$ is true and $0$ otherwise. $\omega$ is a weight to balance the magnitude for cost terms and is set to 2 in this paper.

\vspace{-10pt}
\paragraph{Loss Functions.} Our network is trained via losses based on the bipartite matching results and consists of five parts: a classification loss, a plane parameter loss, a plane center loss, an embedding loss, and a pixel-wise depth loss. The classification loss is defined as:
\begin{equation}
\mathcal{L}_{\text {cls }}^{(i)} = -\log p_{{\hat{\sigma}}(i)}\left( \hat{p}_{i}\right).
\end{equation}
The plane parameter loss is defined as:
\begin{align}
\label{eq:param}
\mathcal{L}_{\text {parm }}^{(i)} =
& \mathbbm{1}_{\{\hat{p}_{i}=1\}}   \,          L_{1}\left(\hat{{n}}_{i}, {n}_{{\hat{\sigma}}(i)}\right) + \nonumber\\
& \mathbbm{1}_{\{\hat{p}_{i}=1\}}   \,  \beta_1 \left(1- \cos\left(\hat{{n}}_{i}, {n}_{{\hat{\sigma}}(i)}\right)\right) + \nonumber\\
& \mathbbm{1}_{\{\hat{p}_{i}=1\}}   \,  \beta_2 \sum\limits_{q\in Q_{i}} \| {n}_{{\hat{\sigma}}(i)}^{T} q - 1 \|,
\end{align}
where $Q_{i}$ is the set of 3D points calculated from pixels belonging to ground truth plane instance $\hat{s}_{p}^{i}$ via the ground truth depth map. $\beta_1 = 5$ and $\beta_2 = 2$ in this paper. 
The embedding loss is defined according to PlaneAE \cite{PlaneAE19} which consists of two terms, called pull loss and push loss. The formulation is as follows:
\begin{equation}
\mathcal{L}_{\text {emb}}^{(i)} = \mathcal{L}_{\text {pull}}^{(i)} + \mathcal{L}_{\text {push}}^{(i)} ,
\end{equation}
where
\begin{equation}
\label{eq:pull}
\mathcal{L}_{\text {pull}}^{(i)} = \mathbbm{1}_{\{\hat{p}_{i}=1\}} \sum\limits_{g\in G_{i}} \max \left( \|{\mathcal{E}}_{{\hat{\sigma}}(i)} - g\|-\delta_1, 0\right),
\end{equation}
\begin{equation}
\label{eq:push}
\small
\mathcal{L}_{\text {push}}^{(i)} = \mathbbm{1}_{\{\hat{p}_{i}=1\}}\sum\limits_{j=1, j\ne i}^{M} \max \left( \delta_2 - \| {\mathcal{E}_{{\hat{\sigma}}(i)}} - {\mathcal{E}_{{\hat{\sigma}}(j)}} \|, 0 \right)
\end{equation}
Here, $G_i$ is a set of embedding vectors of pixels belonging to the ground truth plane instance $\hat{s}_{p}^{i}$. ${\mathcal{E}_{{\hat{\sigma}}(i)}}$ and ${\mathcal{E}_{{\hat{\sigma}}(j)}}$ are the embeddings of predicted plane instances. $\delta_1$ and $\delta_2$ are the margin for pull and push losses and are set to 0.5 and 1.5 in this paper. The depth loss is defined as:
\begin{equation}
\mathcal{L}_{\mathcal{D}} = L_{1} \left( d, \hat{d} \right),
\end{equation}
where $d$ indicates the predicted depth map from Depth Decoder and $\hat{d}$ is the ground truth depth map. The plane center losses of plane instances and pixels can be defined as:
\begin{equation}
\mathcal{L}_{\text {c}}^{(i)} = \mathbbm{1}_{\{\hat{p}_{i}=1\}}  \,  L_{2}\left(\hat{{c}}_{i}, {c}_{{\hat{\sigma}}(i)}\right),
\end{equation}

\begin{equation}
{\mathcal{L}}_{\mathcal{C}} = L_{2}\left( \mathcal{C}, \hat{\mathcal{C}} \right),
\end{equation}
where $\hat{{c}}_{i}$ and $\hat{\mathcal{C}}$ indicate instance-wise and pixel-wise ground truth plane centers.

The final loss is the sum of the above losses and is defined as:
\begin{equation}
\mathcal{L} = 
\sum\limits_{i=1}^{K}\left( \mathcal{L}_{\text {cls }}^{(i)} + \mathcal{L}_{\text {parm }}^{(i)} + \mathcal{L}_{\text{c}}^{(i)} + \lambda\mathcal{L}_{\text {emb}}^{(i)} \right)
+\mathcal{L}_{\mathcal{D}} + {\mathcal{L}}_{\mathcal{C}},
\end{equation}
where $\lambda$ is a weight to balance the magnitude for loss terms and is set to 5 in this paper. Besides, according to DETR \cite{carion2020DETR}, we also predict plane instances from $O_c$ and $O_l$ in the Structure-guided Plane Decoder and serve them as the intermediate supervisions. The weights of these intermediate supervisions are set to 0.1.

\section{Experiments}
\label{sec:experiments}

In this section, we conduct experiments on the public ScanNet dataset \cite{dai2017scannet} and NYUv2 dataset \cite{silberman2012indoor} to evaluate and analyze the performance of the proposed PlaneTR.

\subsection{Dataset and Metrics}
\paragraph{ScanNet.} The ScanNet dataset \cite{dai2017scannet} is a large indoor RGB-D video dataset. To train and test our network, we use the plane data processed by PlaneNet \cite{planeNet} from the original ScanNet dataset. The processed dataset contains 50,000 training and 760 testing images with an image size of $256\times 192$.

\vspace{-10pt}
\paragraph{NYUv2 and its Variant.} The NYUv2 dataset \cite{silberman2012indoor} is an indoor RGB-D dataset. The official NYUv2 dataset contains 795 training and 654 testing images with ground truth depth maps. The image size is $640\times 480$. To evaluate the plane recovery performance, we use a variant of NYUv2 dataset, namely NYUv2-Plane, which is generated by PlaneAE \cite{PlaneAE19} with an image size of $256\times 192$.

\vspace{-10pt}
\paragraph{Evaluation Metrics.} Following PlaneNet \cite{planeNet}, we apply plane and pixel recalls to evaluate the plane detection performance. The plane (pixel) recalls are defined as the percentage of the correctly predicted ground truth plane instances (pixels). A plane is considered as a correctly predicted plane if its Intersection over Union (IOU) is larger than 0.5 and the depth (surface normal) error is less than a threshold. 
Besides, to further evaluate the segmentation performance, we apply three popular metrics used in segmentation \cite{arbelaez2010contour, PlaneRecover} called rand index (RI), variation of information (VI) and segmentation covering (SC). On the NYUv2 dataset, we also evaluate the accuracy of depths inferred from reconstructed planes. Following \cite{eigen2015predicting}, we apply these popular depth metrics: mean absolution relative error ($\text{Rel}$), mean ${\log}_{10}$ error (${\log}_{10}$), root mean square error ($\text{RMSE}$) and the threshold accuracy ($\delta_i < {1.25}^i$, $i=1,2,3$).

\subsection{Implementation Detail}
Our network is implemented with Pytorch~\cite{pytorch} and trained with the Adam optimizer \cite{kingma2014adam}. We train our network on the ScanNet training set with total of 60 epochs on 3 GPUs. The initial learning rate is set to $1\times10^{-4}$ and reduced to half after every 15 epochs. The batch size is set to 24 for each GPU and the weight decay is set to $1\times 10^{-5}$. 

\subsection{Results on the ScanNet Dataset}
\begin{figure*}
\centering
\renewcommand\tabcolsep{15pt}
\begin{tabular}{rccccccc}
\raisebox{20pt}{\rotatebox[origin=c]{90}{Input}}
\includegraphics[width=0.12\linewidth]{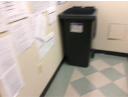}
\includegraphics[width=0.12\linewidth]{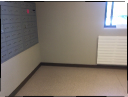}
\includegraphics[width=0.12\linewidth]{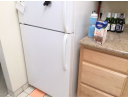}
\includegraphics[width=0.12\linewidth]{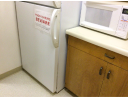}
\includegraphics[width=0.12\linewidth]{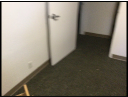}
\includegraphics[width=0.12\linewidth]{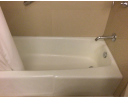}
\includegraphics[width=0.12\linewidth]{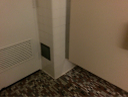}\\
\raisebox{20pt}{\rotatebox[origin=c]{90}{Planes}}
\includegraphics[width=0.12\linewidth]{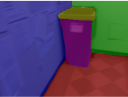}
\includegraphics[width=0.12\linewidth]{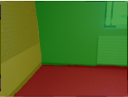}
\includegraphics[width=0.12\linewidth]{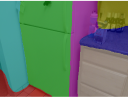}
\includegraphics[width=0.12\linewidth]{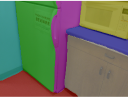}
\includegraphics[width=0.12\linewidth]{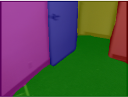}
\includegraphics[width=0.12\linewidth]{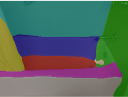}
\includegraphics[width=0.12\linewidth]{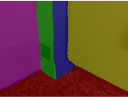}\\
\raisebox{20pt}{\rotatebox[origin=c]{90}{Depth}}
\includegraphics[width=0.12\linewidth]{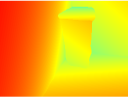}
\includegraphics[width=0.12\linewidth]{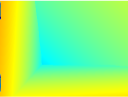}
\includegraphics[width=0.12\linewidth]{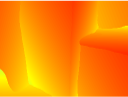}
\includegraphics[width=0.12\linewidth]{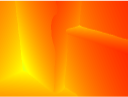}
\includegraphics[width=0.12\linewidth]{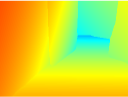}
\includegraphics[width=0.12\linewidth]{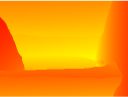}
\includegraphics[width=0.12\linewidth]{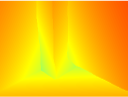}\\
\raisebox{20pt}{\rotatebox[origin=c]{90}{3D Models}}
\includegraphics[width=0.12\linewidth]{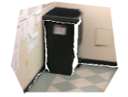}
\includegraphics[width=0.12\linewidth]{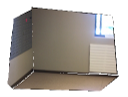}
\includegraphics[width=0.12\linewidth]{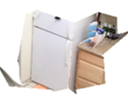}
\includegraphics[width=0.12\linewidth]{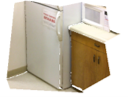}
\includegraphics[width=0.12\linewidth]{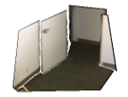}
\includegraphics[width=0.12\linewidth]{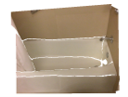}
\includegraphics[width=0.12\linewidth]{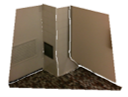}
\end{tabular}  
\caption{3D plane reconstruction results of PlaneTR on the ScanNet dataset.}
\vspace{-2mm}
\label{fig:3DPlane}
\end{figure*}

\begin{figure*}
\centering
\subfigure[{\scriptsize Input}]{
\begin{minipage}[b]{0.115\linewidth}
\includegraphics[width=1\linewidth]{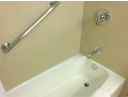}
\includegraphics[width=1\linewidth]{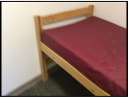}
\includegraphics[width=1\linewidth]{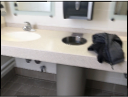}
\includegraphics[width=1\linewidth]{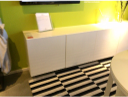}
\includegraphics[width=1\linewidth]{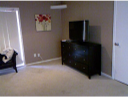}
\includegraphics[width=1\linewidth]{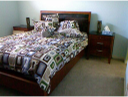}
\end{minipage}}
\subfigure[{\scriptsize Line segments}]{
\begin{minipage}[b]{0.115\linewidth}
\includegraphics[width=1\linewidth]{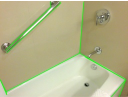}
\includegraphics[width=1\linewidth]{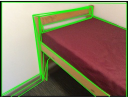}
\includegraphics[width=1\linewidth]{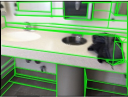}
\includegraphics[width=1\linewidth]{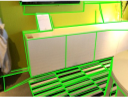}
\includegraphics[width=1\linewidth]{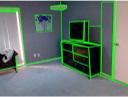}
\includegraphics[width=1\linewidth]{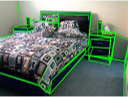}
\end{minipage}
}
\subfigure[{\scriptsize PlaneNet}]{
\begin{minipage}[b]{0.115\linewidth}
\includegraphics[width=1\linewidth]{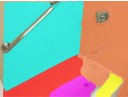}
\includegraphics[width=1\linewidth]{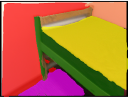}
\includegraphics[width=1\linewidth]{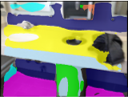}
\includegraphics[width=1\linewidth]{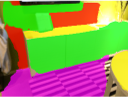}
\includegraphics[width=1\linewidth]{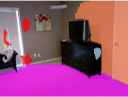}
\includegraphics[width=1\linewidth]{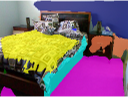}
\end{minipage}
}
\subfigure[{\scriptsize PlaneRCNN}]{
\begin{minipage}[b]{0.115\linewidth}
\includegraphics[width=1\linewidth]{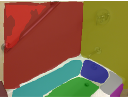}
\includegraphics[width=1\linewidth]{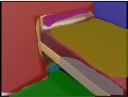}
\includegraphics[width=1\linewidth]{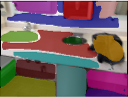}
\includegraphics[width=1\linewidth]{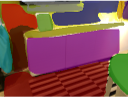}
\includegraphics[width=1\linewidth]{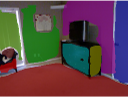}
\includegraphics[width=1\linewidth]{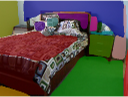}
\end{minipage}
}
\subfigure[{\scriptsize PlaneAE}]{
\begin{minipage}[b]{0.115\linewidth}
\includegraphics[width=1\linewidth]{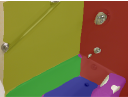}
\includegraphics[width=1\linewidth]{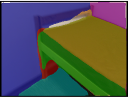}
\includegraphics[width=1\linewidth]{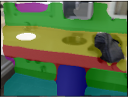}
\includegraphics[width=1\linewidth]{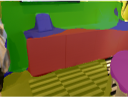}
\includegraphics[width=1\linewidth]{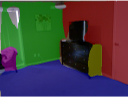}
\includegraphics[width=1\linewidth]{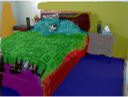}
\end{minipage}
}
\subfigure[{\scriptsize Ours}]{
\begin{minipage}[b]{0.115\linewidth}
\includegraphics[width=1\linewidth]{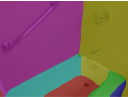}
\includegraphics[width=1\linewidth]{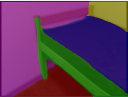}
\includegraphics[width=1\linewidth]{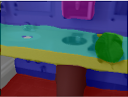}
\includegraphics[width=1\linewidth]{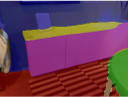}
\includegraphics[width=1\linewidth]{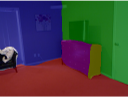}
\includegraphics[width=1\linewidth]{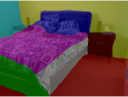}
\end{minipage}}
\subfigure[{\scriptsize Ground Truth}]{
\begin{minipage}[b]{0.115\linewidth}
\includegraphics[width=1\linewidth]{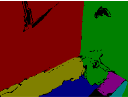}
\includegraphics[width=1\linewidth]{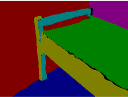}
\includegraphics[width=1\linewidth]{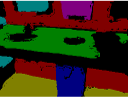}
\includegraphics[width=1\linewidth]{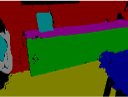}
\includegraphics[width=1\linewidth]{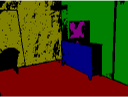}
\includegraphics[width=1\linewidth]{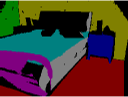}
\end{minipage}}
\caption{Comparison of plane instance segmentation results on the ScanNet dataset (rows: 1-4) and NYUv2-Plane dataset (rows: 5-6).}
\label{fig:plane_compare}
\end{figure*}

\paragraph{Qualitative Results.} Fig.~\ref{fig:3DPlane} shows the plane detection and reconstruction results of our method. PlaneTR can effectively detect and reasonably reconstruct the planar structures in scenes. In the first 4 rows of Fig.~\ref{fig:plane_compare}, we visualize the plane segmentation results of our PlaneTR and the state-of-the-art CNN-based approaches (including PlaneNet \cite{planeNet}, PlaneAE \cite{PlaneAE19} and PlaneRCNN \cite{planercnn}) on the ScanNet dataset~\cite{dai2017scannet}.
PlaneNet and PlaneAE usually incorrectly merge different planes. PlaneRCNN tends to divide a large plane into several plane instances such as the wall and floor shown in the third row of Fig.~\ref{fig:plane_compare}. Besides, the plane segmentation masks detected by these methods are always incomplete. In contrast to those methods, our PlaneTR can correctly detect holistic plane structures with completed segmentation masks.

\begin{figure}
\centering
\includegraphics[scale=0.32]{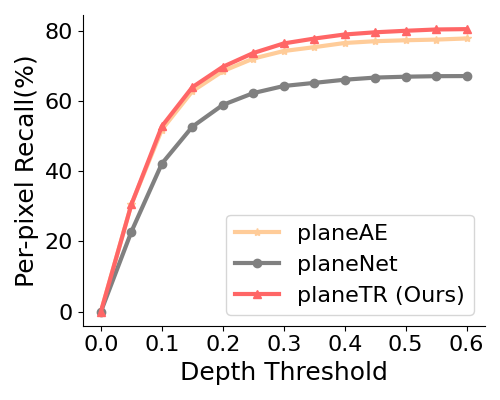} \hspace{-4pt}
\includegraphics[scale=0.32]{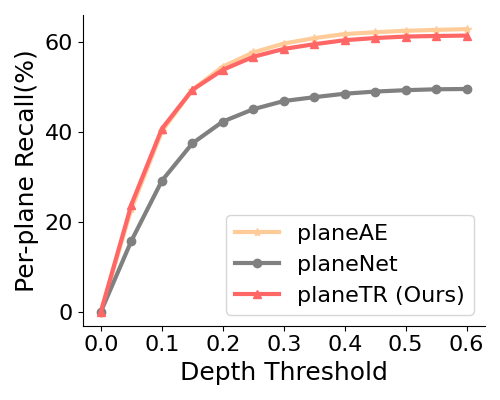}
\caption{Plane and pixel recalls on the ScanNet dataset.}
\label{fig:plane_recall}
\end{figure}

\vspace{-10pt}
\paragraph{Quantitative Results.} We then compare the quantitative performance of our method with PlaneNet \cite{planeNet}, PlaneAE \cite{PlaneAE19} and PlaneRCNN \cite{planercnn}. Note that the PlaneRCNN is trained on a new benchmark constructed from the ScanNetv2 dataset \cite{dai2017scannet} which is different from other methods. Thus, we only treat PlaneRCNN as a reference when evaluating on the ScanNet dataset and mainly compare with the remaining methods. In Table.~\ref{tab:scannet_nyu_seg}, we show the results of plane segmentation performance. Our method significantly outperforms all other methods. In Fig.~\ref{fig:plane_recall}, we show the plane and pixel recalls of various methods and the depth error threshold varies from 0m to 0.6m with an increment of 0.05m. Our method significantly outperforms PlaneNet with all thresholds. When compared with PlaneAE, our method performs better in pixel recalls which are benefited from our completed predicted plane segmentation masks as shown in Fig~\ref{fig:plane_compare}. For the plane recalls, our method performs slightly lower than PlaneAE. It is mainly because that our method tends to detect holistic plane structures which may lead to a missing of some planes such as the white board on the table as shown in the fourth row of Fig.~\ref{fig:plane_compare}.


\begin{table}
\centering
\resizebox{0.95\linewidth}{!}{ 
    \begin{tabular}{c|ccc|ccc}
    \toprule
        \multirow{2}{*}{Method} & \multicolumn{3}{c|}{ScanNet} & \multicolumn{3}{c}{NYUv2-Plane} \\ 
                                & VI~$\downarrow$ & RI~$\uparrow$ & SC~$\uparrow$ & VI~$\downarrow$ & RI~$\uparrow$ & SC~$\uparrow$\\\midrule
        PlaneNet \cite{planeNet}        & 1.259 & 0.858 & 0.716 & 1.813 & 0.753 & 0.558 \\
        PlaneRCNN \cite{planercnn}      & 1.337 & 0.845 & 0.690 & 1.596 & 0.839 & 0.612 \\
        PlaneAE \cite{PlaneAE19}        & 1.025 & 0.907 & 0.791 & 1.393 & 0.887 & 0.681 \\
        Ours                     & \textbf{0.767} & \textbf{0.925} & \textbf{0.838} & \textbf{1.110} & \textbf{0.898} & \textbf{0.726}  \\ \bottomrule
    \end{tabular}
}
\vspace{5pt}  
\caption{Comparison of plane instance segmentation results on the ScanNet dataset and NYUv2-Plane dataset.} 
\label{tab:scannet_nyu_seg}
\end{table}


\subsection{Results on the NYUv2 Dataset}
To verify the generalization of our method, we first evaluate the plane segmentation performance of our method on the NYUv2-Plane dataset. In Table.~\ref{tab:scannet_nyu_seg}, we show the comparison results against PlaneNet \cite{planeNet}, PlaneAE \cite{PlaneAE19}, and PlaneRCNN \cite{planercnn}. Our method outperforms all other methods with a large margin in all metrics. Then, we further evaluate the pixel-wise depth accuracy on the NYUv2 dataset. In this experiment, we achieve the final depth map by first calculate the depth values of plane regions via the predicted 3D planar parameters and then fill out the non-plane region with predicted pixel-wise depth from the Depth Decoder. As shown in Table.~\ref{tab:nyu_depth}, our method outperforms PlaneNet and PlaneAE while PlaneRCNN achieves the best depth performance. Note that PlaneRCNN is trained and tested with the image size of 640$\times$640 which is larger than the image size used in other methods (256$\times$192). Such a large image size is beneficial to PlaneRCNN on the depth performance. The comparison of qualitative plane segmentation results on NYUv2-Plane dataset is shown in the last two rows of Fig.~\ref{fig:plane_compare}. Our method can detect complete and reasonable planes in the scene which are better than other methods.



\begin{table}
\centering
\resizebox{0.95\linewidth}{!}{ 
    \begin{tabular}{c|c|c|c|c}
    \toprule
        Method & PlaneNet & PlaneAE & PlaneRCNN & Ours \\\midrule
        Rel$\downarrow$               & 0.236 & 0.205 & \textbf{0.183} & 0.195 \\ 
        ${\text{log}_{10}}\downarrow$ & 0.124 & 0.097 & \textbf{0.076} & 0.095 \\
        RMSE$\downarrow$              & 0.913 & 0.820 & \textbf{0.619} & 0.803 \\ \midrule
        $\delta_{1}\uparrow$ & 53.0 & 61.3 & \textbf{71.8} & 63.3 \\
        $\delta_{2}\uparrow$ & 78.3 & 87.2 & \textbf{93.1} & 88.2 \\
        $\delta_{3}\uparrow$ & 90.4 & 95.8 & \textbf{98.3} & 96.1 \\ \bottomrule
    \end{tabular}
    }
\vspace{5pt}  
\caption{Depth accuracy comparison on the NYUv2 dataset.}
\label{tab:nyu_depth}
\end{table}



\begin{figure*}[htbp]
\centering
\subfigure[Image]{
\includegraphics[scale=0.25]{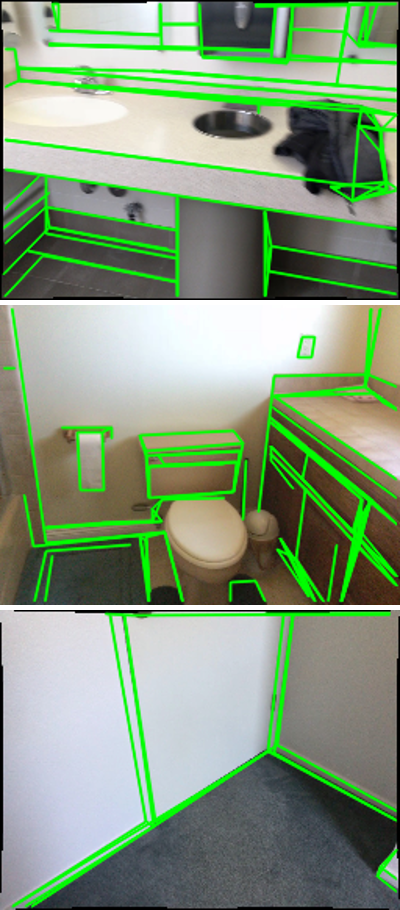}
} 
\subfigure[Seg. Results]{
\includegraphics[scale=0.25]{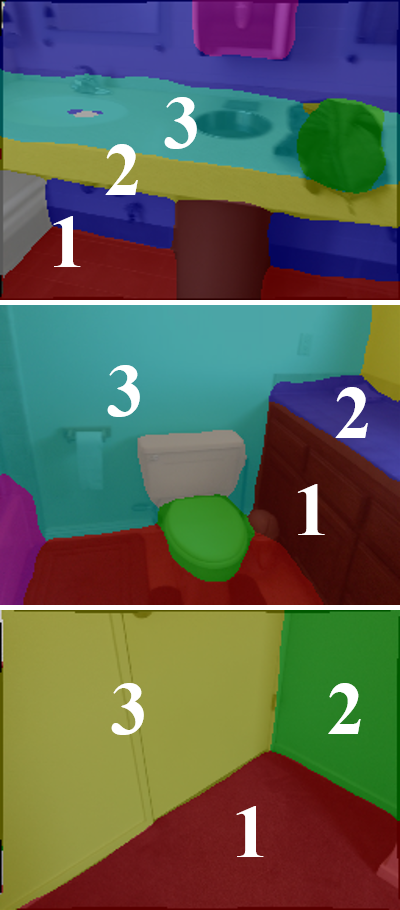}
}
\subfigure[Attention of Plane 1]{
\includegraphics[scale=0.25]{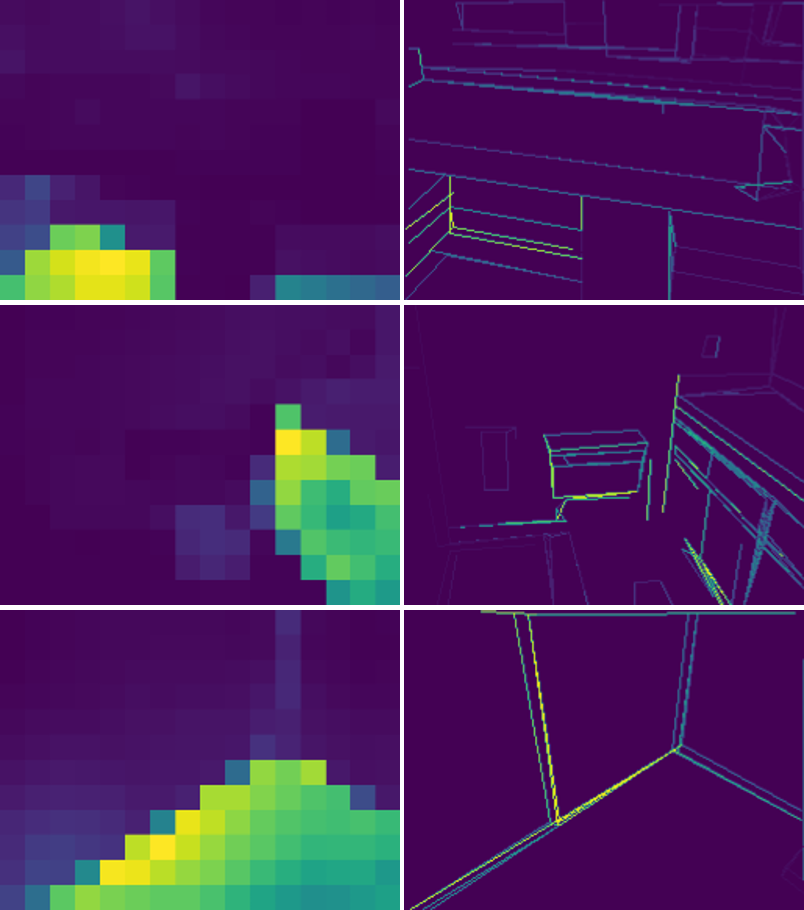}
}
\subfigure[Attention of Plane 2]{
\includegraphics[scale=0.25]{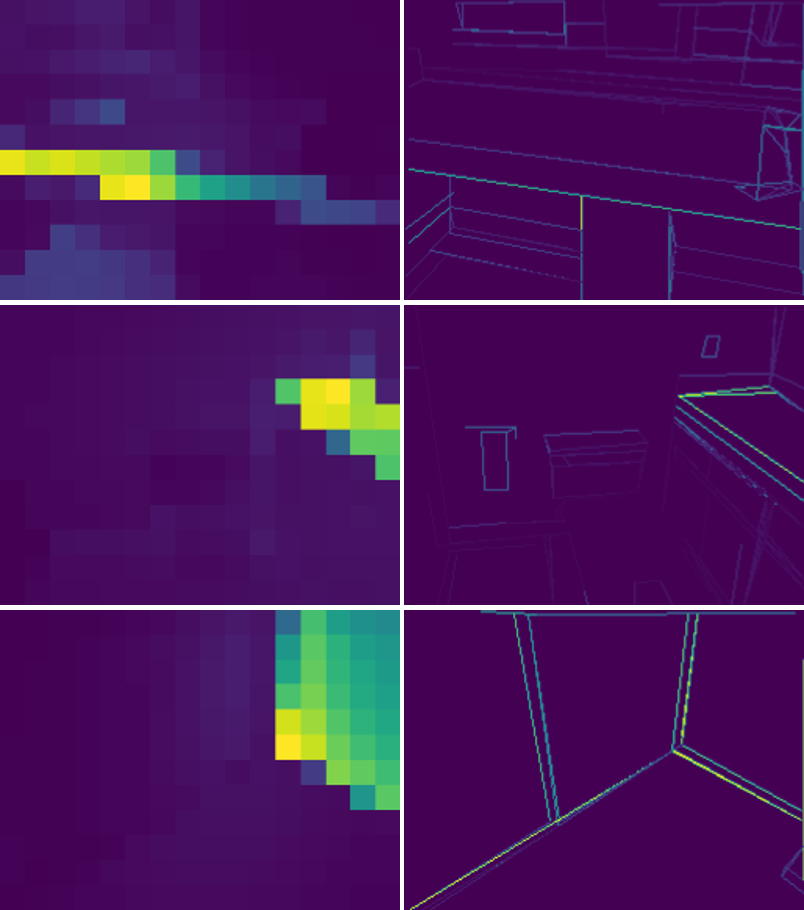}
}
\subfigure[Attention of Plane 3]{
\includegraphics[scale=0.25]{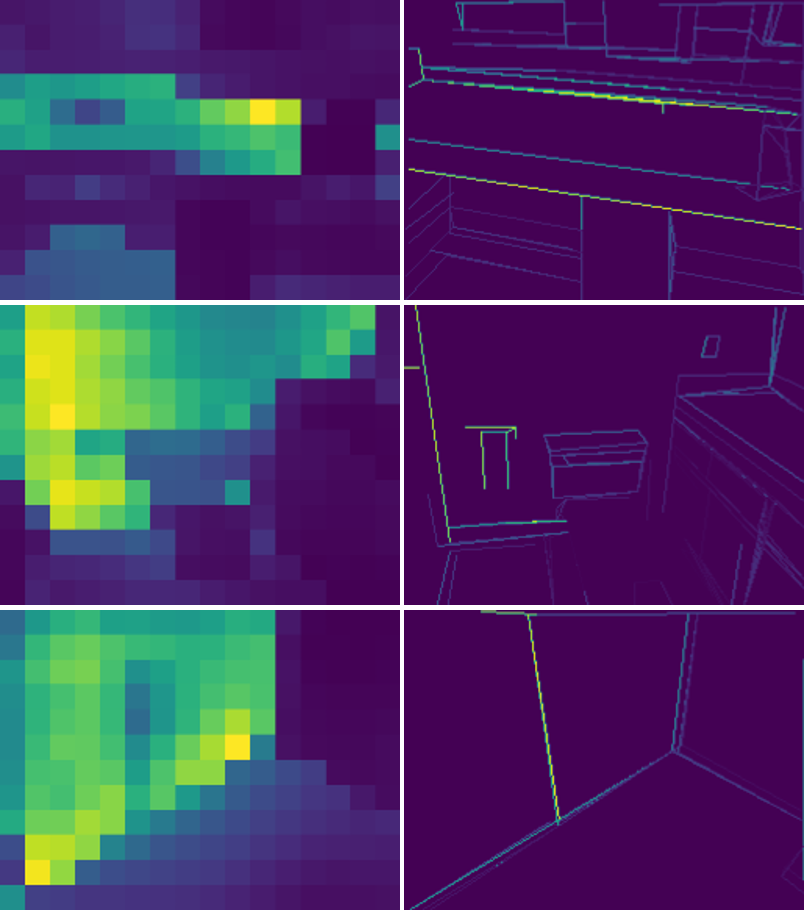}
}
\caption{Context and line segments attention maps in the Structure-guided Plane Decoder.}  
\vspace{-5mm}
\label{fig:att_map}
\end{figure*}

\begin{figure}
\centering
\subfigure[Image]{
\begin{minipage}[b]{0.24\linewidth}
\includegraphics[width=1\linewidth]{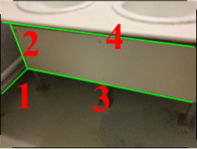}
\end{minipage}} \hspace{-5pt}
\subfigure[w/o line]{
\begin{minipage}[b]{0.24\linewidth}
\includegraphics[width=1\linewidth]{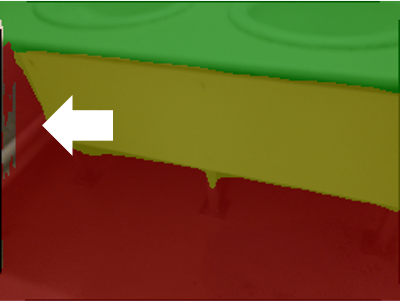}
\end{minipage}} \hspace{-5pt}
\subfigure[w/ line 3 + 4]{
\begin{minipage}[b]{0.24\linewidth}
\includegraphics[width=1\linewidth]{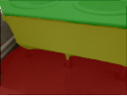}
\end{minipage}} \hspace{-5pt}
\subfigure[w/ line 1 + 2]{
\begin{minipage}[b]{0.24\linewidth}
\includegraphics[width=1\linewidth]{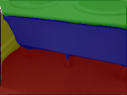}
\end{minipage}} \vspace{-8pt}

\subfigure[Image]{
\begin{minipage}[b]{0.24\linewidth}
\includegraphics[width=1\linewidth]{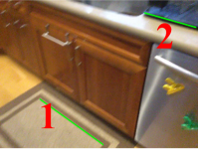}
\end{minipage}} \hspace{-5pt}
\subfigure[w/o line]{
\begin{minipage}[b]{0.24\linewidth}
\includegraphics[width=1\linewidth]{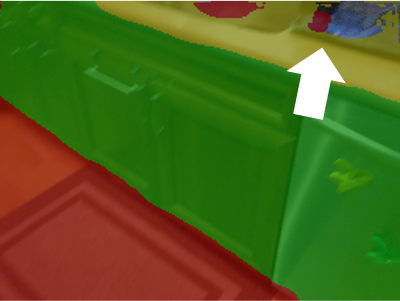}
\end{minipage}} \hspace{-5pt}
\subfigure[w/ line 1]{
\begin{minipage}[b]{0.24\linewidth}
\includegraphics[width=1\linewidth]{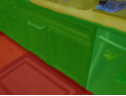}
\end{minipage}} \hspace{-5pt}
\subfigure[w/ line 2]{
\begin{minipage}[b]{0.24\linewidth}
\includegraphics[width=1\linewidth]{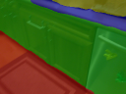}
\end{minipage}} 
\caption{Illustration of the plane detection guided by line segments in PlaneTR.} 
\label{fig:DiffLines}
\end{figure}

\begin{table}
\centering
\resizebox{1.0\linewidth}{!}{
    \begin{tabular}{cc|cccc|ccc}
    \toprule
        \multicolumn{2}{c|}{Settings} & \multicolumn{4}{c|}{Per-plane recall (depth $\&$ normal)} & \multicolumn{3}{c}{Plane Segmentation}  \\ 
        \makecell{Line\\Segments} & \makecell{Plane\\Center} & @0.10 m& @0.60 m& @$5^{\circ}$ & @$30^{\circ}$ & VI~$\downarrow$ & RI~$\uparrow$ & SC~$\uparrow$ \\\midrule
                      &              & 35.71 & 57.03 & 38.77 & 56.40 & 0.885 & 0.913 & 0.809 \\
                      & $\checkmark$ & 38.07 & 60.33 & 40.85 & 59.81 & 0.778 & 0.923 & 0.833 \\
         $\checkmark$ &              & 38.44 & 58.82 & 41.72 & 58.36 & 0.831 & 0.919 & 0.823 \\
         $\checkmark$ & $\checkmark$ & \textbf{40.74} & \textbf{61.49} & \textbf{43.14} & \textbf{60.68} & \textbf{0.767} & \textbf{0.925} & \textbf{0.838} \\\bottomrule
    \end{tabular}
}
\vspace{5pt}
\caption{Ablation studies of proposed PlaneTR on the ScanNet dataset.} 
\label{tab:ablation}
\end{table}

\subsection{Ablation Studies}
In this section, we conduct experiments on the ScanNet dataset to validate various components in PlaneTR. We evaluate the performance by using plane recalls based on depth and normal thresholds, respectively. Specifically, the depth thresholds are set to 0.1m and 0.6m. The normal thresholds are set to $5^{\circ}$ and $30^{\circ}$. Besides, we also use VI, RI, and SC to evaluate the plane segmentation performance.

\paragraph{Auxiliary Plane Center Prediction.} We first validate the effectiveness of auxiliary plane center prediction which we used to assist the learning of our method. As shown in Table.~\ref{tab:ablation}, learning extra plane centers during training is beneficial for our network to achieve better results. One main reason is that plane centers provide important location information of planes which can help the plane queries to learn where the planes are in an image.

\vspace{-10pt}
\paragraph{Line Segments.} Then we further validate the effectiveness of line segments used in our Structure-guided Plane Decoder. In the setting without line segments, the plane decoder is trained by only taking the context sequence as input to predict plane instances. As shown in Table.~\ref{tab:ablation}, we can observe that both plane recalls and plane segmentation results are improved by applying line segments. In Fig.~\ref{fig:DiffLines}, we list two examples to further show how the input line segments guide the detection of planes based on our final model. The setting `w/o line' means we input an empty line sequence into the network. By inputting relative line segments (e.g. line 1 and line 2 in the first example) into the network, the missed plane can be successfully detected. It demonstrates that our Structure-guided Plane Decoder has learned the relationship between planes and line segments.

\begin{table}
\centering
\resizebox{0.85\linewidth}{!}{
    \begin{tabular}{c|c|c|c|c}
    \toprule
        \multirow{2}{*}{Number} & \multicolumn{4}{c}{Per-plane recall}\\ 
                                & @0.10 m & @0.60 m & @$5^{\circ}$ & @$30^{\circ}$ \\\midrule
        20                      & \textbf{40.74} & \textbf{61.49} & 43.14 & \textbf{60.68} \\
        30                      & 40.58 & 61.08 & \textbf{43.36} & 60.35 \\
        40                      & 39.25 & 60.11 & 42.51 & 59.41 \\\bottomrule
    \end{tabular}
    }
\vspace{5pt}
\caption{Plane recalls of PlaneTR with different numbers of plane queries on the ScanNet dataset.} 
\label{tab:plane_num}
\end{table}

\vspace{-10pt}
\paragraph{Plane Query Number.} We also experiment the influence of plane query numbers and the results are shown in Table.~\ref{tab:plane_num}. We can observe that as the number of plane queries increases from 20 to 40, the plane recalls have been degraded to some extent. One reasonable explanation is that the plane number of each image in the ScanNet dataset is lower than 20 and a large number of plane queries will lead to an imbalance between positive and negative samples which may degrade the network performance.

\vspace{-10pt}
\paragraph{Attention Visualization.}
In Fig.~\ref{fig:att_map}, we show attention maps between the predicted plane instances and the input context sequence (line segment sequence) in the Structure-guided Plane Decoder. As we can see, for one plane instance such as the Plane 1 in the second row of Fig.~\ref{fig:att_map}, its attention with the context sequence mainly focuses on those pixels which are located on the plane region. By contrast, its attention with the input line segments shows more complex relationships and global structure information of the scene.

\section{Conclusion}
    
    In this paper, we present a novel model of Transformers, PlaneTR, that simultaneously leverages the context information and global structure cues to recover 3D planes from a single image with state-of-the-art performance obtained. Our PlaneTR represents context features and line segments with tokenized sequences instead of dense maps to address the plane recovery problem in a sequence-to-sequence manner. Different from the existing CNNs-based plane recovery or structure-guided learning methods, such a tokenized representation of line segments enables our network to explicitly exploit the holistic structure cues in the end task. 
    In our future work, we plan to study the utilizing of geometric structures in other geometric vision tasks.
    
{\small
\bibliographystyle{ieee_fullname}
\bibliography{egbib}
}

\end{document}